\documentclass[lettersize,journal]{IEEEtran}
\usepackage{amsmath,amsfonts}
\usepackage{algorithmic}
\usepackage{algorithm}
\usepackage{array}
\usepackage{textcomp}
\usepackage{stfloats}
\usepackage{url}
\usepackage{verbatim}
\usepackage{graphicx}
\usepackage{cite}
\usepackage{caption}
\usepackage{subcaption}

\usepackage[T1]{fontenc}
\usepackage[table,xcdraw]{xcolor}

\usepackage{wrapfig}          
\definecolor{blue}{RGB}{0,114,189}
\definecolor{orange}{RGB}{217,83,25}
\definecolor{green}{RGB}{89,142,28}

\definecolor{lightblue}{RGB}{161,227,249}

\definecolor{ForestGreen}{rgb}{0.13, 0.55, 0.13}

\usepackage{multirow}

\usepackage{pifont}
\usepackage{siunitx}
\usepackage{booktabs}
\usepackage{makecell}
\usepackage{amssymb}
\usepackage{pifont}
\newcommand{\xmark}{\ding{55}} 
\usepackage{xcolor}

\DeclareRobustCommand{\redx}{\textcolor{red}{$\times$}}
\DeclareRobustCommand{\redq}{\textcolor{red}{?}} 

\begin{document}

\title{ChatENV: An Interactive Vision-Language Model for Sensor-Guided Environmental Monitoring and Scenario Simulation}

\author{Hosam Elgendy, Ahmed Sharshar, Ahmed Aboeitta, and Mohsen Guizani \\%
Mohamed bin Zayed University of Artificial Intelligence, Abu Dhabi, UAE
\thanks{All authors are with the Mohamed bin Zayed University of Artificial Intelligence, Abu Dhabi, UAE. Correspondence: hosam.elgendy@mbzuai.ac.ae.}}





\maketitle

\begin{abstract}
Understanding environmental changes from remote sensing imagery is vital for climate resilience, urban planning, and ecosystem monitoring. Yet, current vision language models (VLMs) overlook causal signals from environmental sensors, rely on single-source captions prone to stylistic bias, and lack interactive scenario-based reasoning.
 We present \textit{ChatENV}, the first interactive VLM that jointly reasons over satellite image pairs and real-world sensor data. Our framework: (i) creates a 177k-image dataset forming 152k temporal pairs across 62 land-use classes in 197 countries with rich sensor metadata (e.g., temperature, PM\textsubscript{10}, CO); (ii) annotates data using GPT-4o and Gemini 2.0 for stylistic and semantic diversity; and (iii) fine-tunes Qwen-2.5-VL using efficient Low-Rank Adaptation (LoRA) adapters for chat purposes. ChatENV achieves strong performance in temporal and “what-if” reasoning (e.g., BERT-F1 0.902) and rivals or outperforms state-of-the-art temporal models, while supporting interactive scenario-based analysis. This positions ChatENV as a powerful tool for grounded, sensor-aware environmental monitoring.\footnote{The annotated data and code are available on github through: https://github.com/HosamGen/ChatENV/} 
\end{abstract}


\section{Introduction}
Monitoring and understanding environmental dynamics are top priorities for addressing pressing issues like climate change, urban growth, and habitat degradation. Remote sensing, particularly through the use of aerial and satellite imaging, offers extensible monitoring of Earth's surface, enabling real-time detection of changes in land cover, vegetation, urban areas, and other environmental indicators \cite{wulder2012landsat, corradino2019lava}. Recently, Vision-Language Models (VLMs) have come into the spotlight for fusing visual and text data, enabling enhanced automation of complex tasks in remote sensing, ranging from change detection, captioning, and multimodal reasoning \cite{liu2024changeagent, zhang2025georsmllm}. In spite of this, several key challenges remain:

(1) \textbf{Lack of environmental context}, as most VLMs rely solely on images and ignore essential sensor data crucial for understanding and observing geographical changes \cite{liu2024changeagent}; (2) \textbf{Single-source annotation bias}, where using a single language model for data annotation limits linguistic diversity and introduces bias \cite{wang2023skyscript}; (3) \textbf{Limited spatial and semantic diversity}, since widely used datasets like LEVIR-CD \cite{chen2020levircd} and RSICap \cite{hu2025rsgpt} lack the geographic scope and object-class variety necessary for robust modeling \cite{lu2017exploring}; and (4) \textbf{Lack of interactive reasoning}, with current VLMs offering no "what-if" interfaces to support planning and hypothetical decision-making \cite{liu2024changeagent}.

These challenges reflect two open problems in sensor-guided environmental monitoring with vision-language interaction: 
(i) \emph{grounding} remote sensing observations with heterogeneous environmental measurements that are noisy, intermittently missing, and sampled at different temporal granularities; and 
(ii) enabling \emph{interactive} multi-turn reasoning that can refine hypotheses and support scenario-oriented questions (e.g., planning or mitigation) rather than single-shot captions. 
Bridging these gaps requires models that jointly reason over imagery, time, and environmental context while maintaining conversational consistency across turns.

To address these challenges, we introduce \textbf{ChatENV}, a unified framework that combines remote sensing images and real-world environmental sensor data. This combination improves the ability to detect and explain changes over time and space. At the core of our approach is a large-scale satellite-aerial dataset that covers a wide geographic area, includes diverse land types and object categories, and pairs each image (when available) with real-world environmental measurements (e.g., weather and air-quality variables). 
\noindent\textbf{Our contributions are:}

\begin{itemize}

\item \textbf{SOTA Multimodal Dataset.} We release the largest dataset of its kind, containing over \textit{177k} satellite images, divided into \textit{152k} time-separated image pairs, across \textit{62 object classes} and \textit{197 countries}. Each pair is accompanied by sensor-tagged prompts aligned to capture time and rich change captions generated by a dual-model pipeline (GPT-4o and Gemini 2.0), to reduce stylistic bias and increase linguistic diversity. All test captions are manually reviewed to ensure accuracy, providing a strong benchmark for environmental change detection tasks.

\item \textbf{Sensor-Guided Multimodal Learning.} Through the fusion of high-resolution images and temperature, humidity, wind speed, UV index, and emissions data ($PM_{10}$, CO, $NO_2$), our model goes beyond visual changes to include environmental context, leading to a deeper understanding of the underlying causes of change.

\item \textbf{Interactive Vision-Language Chat Model.} ChatENV is a multi-tasking vision-language dialogue model, trained with a multi-turn prompting format and scenario-based supervision, that can describe single images, identify and explain changes between image pairs, and respond to scenario-based questions (e.g., “What if more trees were planted?”). This supports exploring real and imagined changes for environment-informed decision-making.

\end{itemize}

\noindent\textbf{Novelty:}
ChatENV introduces a unified, sensor-conditioned vision--language framework for environmental monitoring that jointly leverages (i) time-separated remote sensing imagery, (ii) real-world environmental measurements aligned to each image timestamp, and (iii) multi-turn, scenario-style supervision. This combination enables a dialogue-capable model to describe scenes, explain temporal changes, and answer \emph{what-if} queries in a single interface while remaining grounded in both visual evidence and contextual environmental signals.

\begin{figure*}
    \centering
    \includegraphics[width=0.92\textwidth]{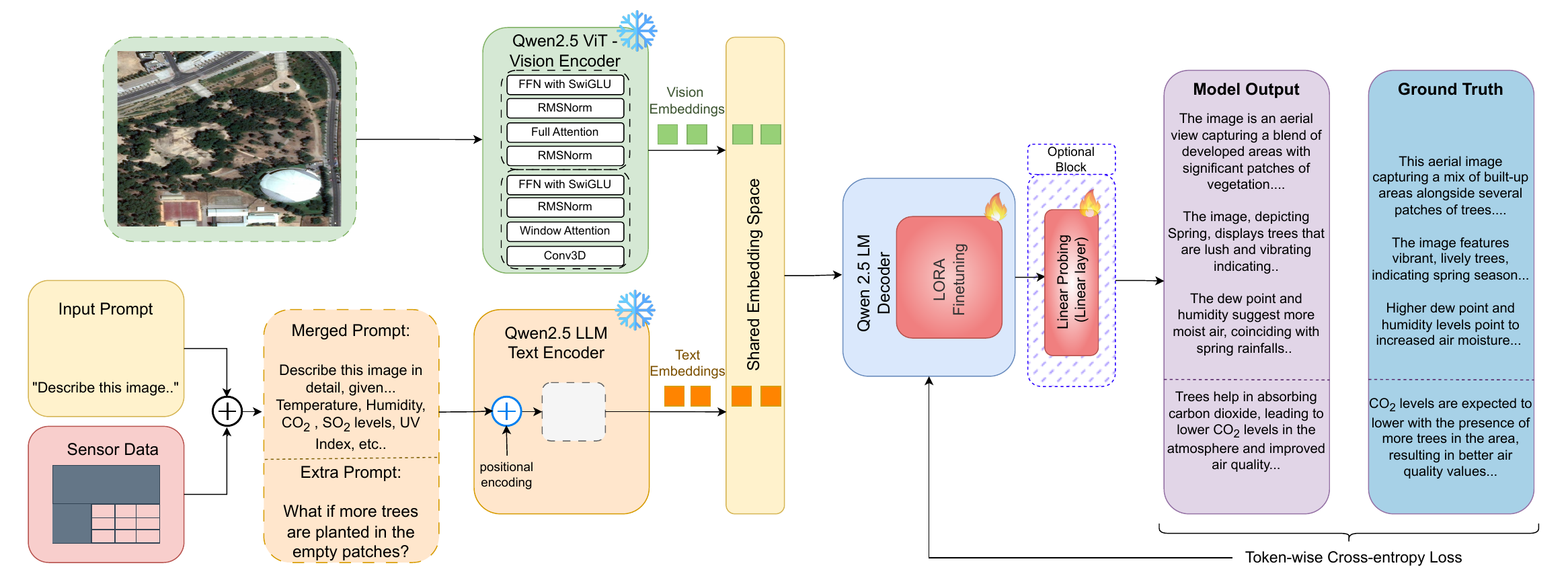}
    \caption{
\textbf{Pipeline overview for \textsc{ChatENV}.}  
Aerial RGB tiles and sensor-tagged prompts (e.g., temperature, humidity, \textsc{CO\textsubscript{2}}) are encoded via \textit{frozen} Qwen 2.5 ViT and text encoders, respectively.  
Their embeddings are projected into a shared space to condition a Qwen 2.5 decoder, with only \textit{LoRA adapters} and an optional \textit{linear probe} trained.  
Token-level cross-entropy on enriched captions trains the model to (i) describe scenes, (ii) reason over current environmental data, and (iii) answer “what-if” queries.}
    \label{fig:overview}
\end{figure*}

\section{Related Work}

\textbf{Benchmark Datasets:}
Several datasets support geospatial vision–language tasks. For captioning, RSICD \cite{lu2017exploring}, NWPU-Captions \cite{Cheng2022}, and RSICap \cite{hu2025rsgpt} offer image-text pairs at varying scales. RSVQA \cite{Lobry2020} and BigEarthNet-based extensions \cite{Yuan2022} focus on spatial Visual Question Answering (VQA), while LEVIR-CD \cite{Chen2020} and LEVIR-CC \cite{Liu2022} provide Change Detection (CD) labels and natural language descriptions. ChangeCLIP \cite{Dong2024} benchmarks semantic CD, and SpaceNet \cite{spacenet} provides long-term urban imagery but lacks text annotations. However, most existing datasets lack temporal variety, environmental sensor data, or support for open-ended user queries. There remains a need for benchmarks that support contextual understanding, multimodal learning, and interactive reasoning.

\textbf{Remote Sensing Foundation Models.}
Remote sensing foundation models such as SpectralGPT~\cite{spectralgpt
}, SeaMo~\cite{seamo}, and FlexiMo~\cite{fleximo} learn general-purpose representations from large-scale Earth observation data, but they are not designed for interactive vision--language dialogue and do not explicitly support in-situ environmental sensor grounding or counterfactual \emph{what-if} querying.

\textbf{Vision{-}Language Models:}
Recent VLMs such as Pixtral \cite{pixtral2024}, DeepSeek-VL \cite{lu2024deepseekvl}, Janus Pro \cite{chen2025janus}, Phi-4-Multimodal \cite{abouelenin2025phi4}, and LLaMA-3-Vision \cite{grattafiori2024llama3} demonstrate strong performance on tasks like captioning, visual reasoning, and instruction following. These models integrate vision encoders directly into Large Language Models (LLMs) and are trained on large-scale image–text datasets. While not designed for remote sensing, their scalability and reasoning capabilities make them promising candidates for domain adaptation.

\textbf{VLMs for Remote Sensing:}
Domain-specific adaptations of VLMs have emerged to tackle geospatial tasks. For example, RSGPT \cite{hu2025rsgpt} fine-tunes Vicuna on RSICap for aerial captions. RS-LLaVA~\cite{bazi2024rsllava} and EarthGPT \cite{Zhang2024} use CLIP/ViT backbones with LLaMA-based LLMs for captioning and VQA, incorporating multi-sensor data. RemoteCLIP~\cite{remoteclip} and GeoChat \cite{kuckreja2024geochat} build on large image–text corpora but lack temporal modeling. GeoLLaVA \cite{elgendy2024geollava} uses video-based fine-tuning for change detection, though it still lacks annotation diversity and interactivity.

\textbf{Captioning and VQA in Aerial Imagery:}
Initial work on remote sensing VQA began with RSVQA \cite{Lobry2020}, which used OpenStreetMap for spatial questions, and was extended by \cite{Yuan2022} using BigEarthNet. These efforts rely on synthetic prompts and lack environmental data. In captioning, RSICD introduced a CNN--RNN pipeline, with NWPU-Captions adding attention mechanisms, and more recent work \cite{Bourcier2024} exploring metadata-aware pretraining but is still limited to static imagery.


\begin{table}[t]
    \centering
    \caption{Comparison of Vision-Language Models for Remote Sensing Tasks. 
    Based on sensor inputs, conversational and temporal capabilities, and predictive capacity.}
    \label{tab:vlm_comparison}
    \renewcommand{\arraystretch}{1.1}
    \resizebox{0.48\textwidth}{!}{%
    \begin{tabular}{@{}lcccc@{}}
        \toprule
        \textbf{Model} & \textbf{Interactive} & \textbf{Temporal} & \textbf{What-if} & \textbf{Sensor}\\
        \midrule
        RSGPT~\cite{hu2025rsgpt}        & \xmark & \xmark & \xmark & \xmark \\
        RS-LLaVA~\cite{Bazi2024}        & \checkmark & \xmark & \xmark & \xmark \\
        EarthGPT~\cite{Zhang2024}       & \checkmark & \xmark & \xmark & \xmark \\
        RemoteCLIP~\cite{remoteclip}    & \xmark & \xmark & \xmark & \xmark \\
        SpectralGPT~\cite{spectralgpt}    & \xmark & \checkmark & \xmark & \xmark \\
        SeaMo~\cite{seamo}    & \xmark & \checkmark & \xmark & \xmark \\
        FlexiMo~\cite{fleximo}    & \xmark & \xmark & \xmark & \xmark \\
        GeoChat~\cite{kuckreja2024geochat} & \checkmark & \xmark & \xmark & \xmark \\
        GeoLLaVA~\cite{elgendy2024geollava} & \checkmark & \checkmark & \xmark & \xmark \\
        Tree-GPT~\cite{Du2023}          & \checkmark & \xmark & \xmark & \xmark \\
        ChangeCLIP~\cite{Dong2024}      & \xmark & \checkmark & \xmark & \xmark \\
        CD-VQA~\cite{Yuan2022}          & \xmark & \checkmark & \xmark & \xmark \\
        TEOChat~\cite{irvin2024teochat} & \checkmark & \checkmark & \checkmark & \xmark \\
        \rowcolor{lightblue}
        \textbf{Ours: ChatENV}          & \checkmark & \checkmark & \checkmark & \checkmark \\
        \bottomrule
    \end{tabular}}
    \vspace{-5pt}
\end{table}

\textbf{Interactive Reasoning for Environmental Monitoring:}  
Conversational VLMs like, RS-LLaVA, and EarthGPT enable multi-turn Q\&A but lack predictive reasoning. Tree-GPT \cite{Du2023} adds geospatial tools for command-based interaction but cannot simulate hypothetical, counterfactual scenarios. Finally, TEOChat \cite{irvin2024teochat} supports scene monitoring by asking questions about objects that change in an image or disasters affecting a location, but it does not incorporate environmental reasoning based on temperature or other meteorological metrics. 

Most prior work either (i) handles \textit{static} imagery only (lacking temporal information), 
(ii) omits \textit{continuous sensor signals}, (iii) uses \textit{templated queries}, or 
(iv) lacks \textit{counterfactual reasoning}. \textit{ChatENV} addresses all four by integrating images and sensor data for grounded, interactive, and predictive environmental understanding, as summarized in Table~\ref{tab:vlm_comparison}.

\section{Data Collection and Pre-processing} \label{sec:dataset} 
\begin{figure*}[t]
    \centering
    \includegraphics[width=0.7\textwidth]{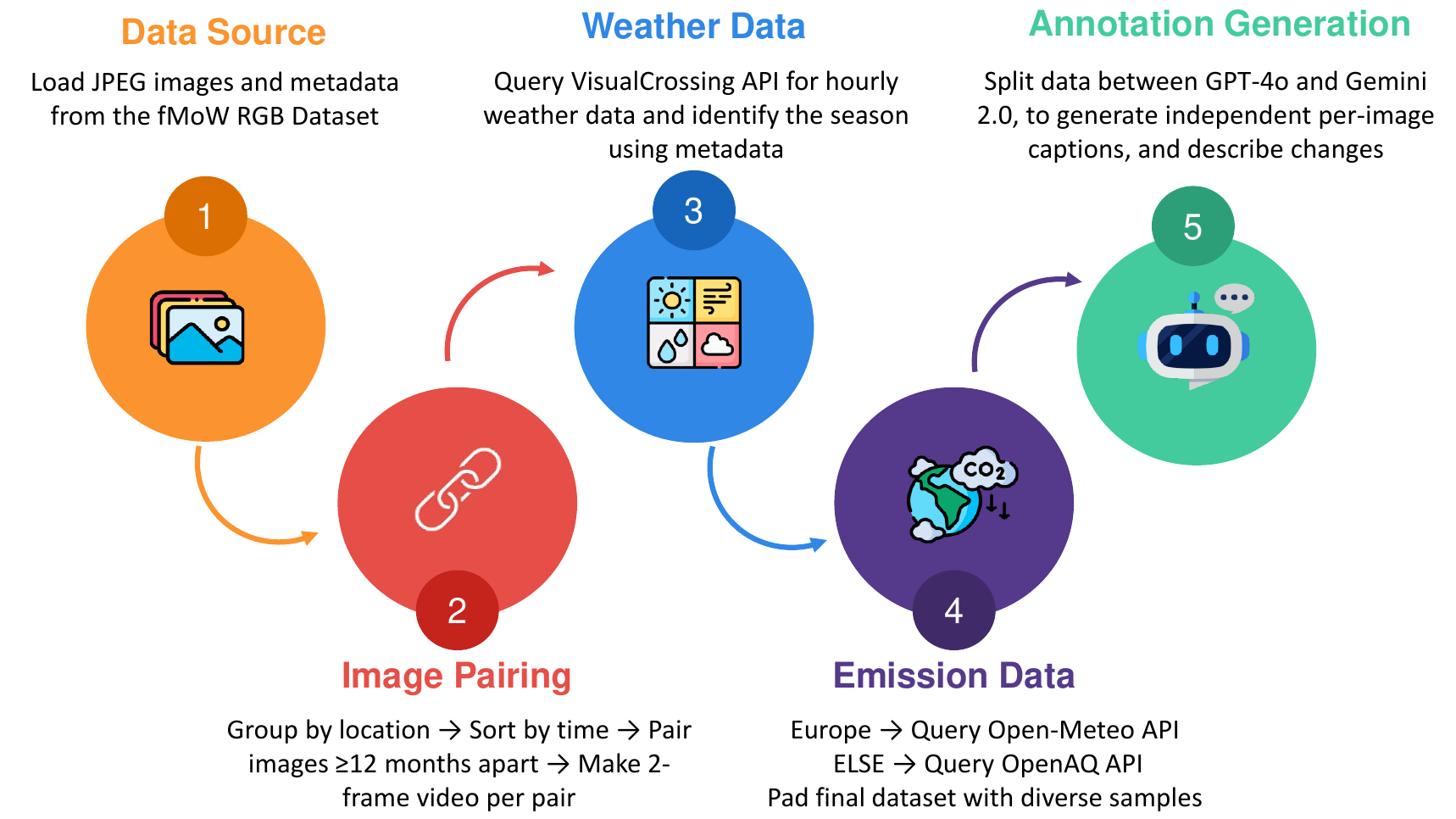}
    \caption{Visual overview of the preprocessing pipeline for environmental change analysis. The process starts with satellite imagery sourcing (Data Source), followed by temporal pairing (Image Pairing), integration of Weather Data, enrichment with emissions (Emission Data), and annotation via GPT-4o and Gemini 2.0 (Annotation Generation).}
    \label{fig:data_collection}
    \vspace{-10pt} 
\end{figure*}

We built a multi-step pipeline to fetch and preprocess high-resolution satellite images from the fMoW dataset, summarized in Figure~\ref{fig:data_collection}.

\textbf{Construct Temporal Pairs from fMoW Dataset.} fMoW RGB dataset~\cite{fmow} consists of $363{,}571$ train images and $53{,}041$ validation images covering years $2002$ to $2017$ globally. With spatial resolution around $0.3$ $m$, it provides increased accuracy for fine-grained change detection between $62$ classes~\cite{satmae} over 197 countries illustrated in Figure~\ref{fig:treemap}. We choose fMoW due to its global coverage and rich spatiotemporal metadata (timestamp and geolocation), which are essential for pairing images over time and aligning each capture with environmental context. We pair images based on location while maintaining a minimum temporal gap of $12$ months. This interval is a dataset-construction parameter (not a model hyperparameter) that follows the same selection as \cite{elgendy2024geollava} to ensure meaningful land-use or structural changes rather than short-term appearance variability. This choice does not alter the model architecture or training procedure. After filtering and removing duplicates, the final training set contains $177k$ images, corresponding to $152k$ pairs, and the test set contains $14k$ images, corresponding to $10k$ pairs. Table~\ref{tab:dataset} provides an overview of dataset splits using the 12-month interval, we release timestamps for all images to allow re-pairing data using alternative gaps (e.g., 6-18 months) depending on the target change scale.

\paragraph{Weather and Emissions Data API} We used fMoW Dataset metadata, containing geospatial locations, country codes, time zones, and timestamps, to fetch weather and atmospheric emissions data through third-party APIs. For weather features, the VisualCrossing API~\cite{visualcrossing} fetched a large set of fields, like \texttt{temperature}, \texttt{dew point}, \texttt{humidity}, and more, by using the capture location and timestamp of each image. Likewise, emissions data were collected using Open-Meteo~\cite{openmeteo} for European samples and OpenAQ~\cite{openaq} for non-European samples, yielding parameters like \texttt{$pm2.5$}, \texttt{$CO$}, \texttt{$NO_2$}, and \texttt{ozone}. We focused on features that appeared most frequently and for which our analyses would be most relevant, and cross-referenced country codes from the fMoW dataset with the country names returned by the APIs to ensure temporal alignment and season-specific context. Features with limited relevance or significant missing data were discarded, while those lacking emissions data were retained to preserve category and geographic diversity. This fusion-mapping of environmental metadata and the dataset’s global, multi-year coverage allowed us to capture temporal, climatic, and atmospheric changes with minimal missing information, supporting downstream modeling work.

\begin{table*}[]
\caption{Comparison between the original fMoW dataset and the modified version with environmental data.}
\resizebox{\textwidth}{!}{%
\begin{tabular}{lcccc}
\hline
                          & \textbf{\# Images (train/test)} & \textbf{\# Annotated Pairs (train/test)} & \textbf{Weather Sensors}  & \textbf{Emissions Sensors}                        \\ \hline
\textbf{fMoW Dataset}     & 363,571 / 53,041                & 0                                        & None                      & None                                              \\
\textbf{Modified Dataset} & 177,719 / 14,806                & 152,710 / 10,020                         & Temp, Humidity, Dew, etc. & CO, O\textsubscript{3}, NO\textsubscript{2}, etc. \\ \hline
\end{tabular}%
}
\label{tab:dataset}
\end{table*}

\paragraph{Annotations Generation} We produced descriptive annotations to capture the most important features of each image and summarize temporal variation between matched samples. Two SOTA LLMs, GPT-4o and Gemini 2.0 Flash Thinking, were used to extend generalization, mitigate stylistic bias, and increase robustness. All prompts included season, semantic category, and environmental sensor data, following a consistent prompting strategy for each model. The dataset was divided equally between ChatGPT and Gemini, with images from the same location assigned to the same model to enforce locality-consistent partitioning. The training set contains 177{,}719 images (152{,}710 pairs) and the test set 14{,}806 images (10{,}020 pairs), each augmented with weather data and text annotations. Of these, 93{,}834 training images and 14{,}322 validation images also contain emissions data, enabling end-to-end analysis of atmospheric factors.

\begin{figure}
    \centering
    \includegraphics[width=\linewidth]{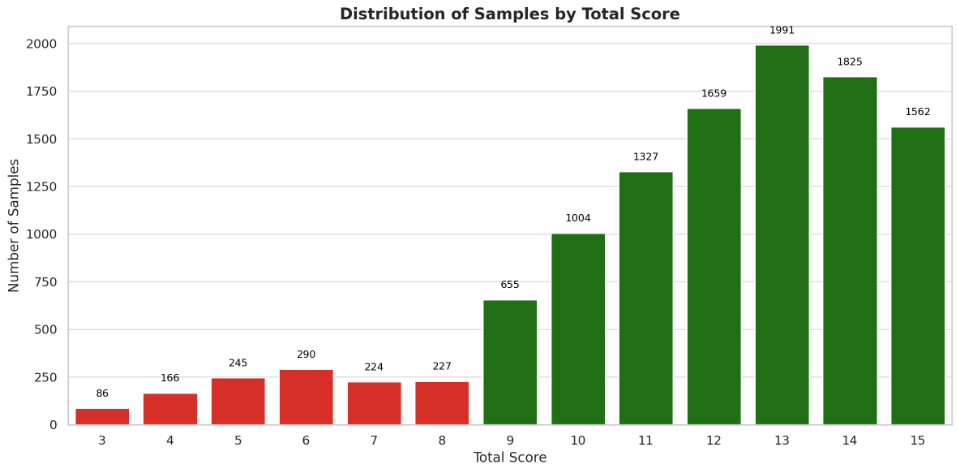}
    \caption{Total distribution of samples by score through manual evaluation of the testing set. For each sample, a rating of 1-5 was given over three criteria. Samples with a total score over 9 points were kept as the testing set.}
    \label{fig:distribution}
\end{figure}

\paragraph{Annotations Manual Evaluation}
Before conducting the model evaluation, the test dataset underwent a thorough manual inspection to ensure the reliability and correctness of the generated annotations of both GPT-4 and Gemini 2.0 models. First, annotators (who were also the authors) verified whether the dominant object in each image aligned with the ground truth class label provided in the original dataset, offering an initial layer of validation to the annotation quality. Second, the annotators were asked targeted questions for each sample, aimed at assessing the accuracy of the generated descriptions with respect to key visual elements and prominent objects present in the image. Each annotator was asked to rate each description on a scale of 1 to 5, over three questions. For each description, the points are summed up, and a threshold of 9 points was used to filter annotations that make it to the test set versus the ones that do not, as elaborated in Figure \ref{fig:distribution}.  To account for the inherent subjectivity involved in evaluating such open-ended descriptions, the dataset was split among the annotators in a non-overlapping manner, thereby avoiding inter-rater discrepancies.

\begin{figure*}[t!]
    \centering
    \includegraphics[width=0.8\textwidth]{Images/treemap_figure.pdf}
    \caption{Treemap visualization showing the distribution of satellite image counts by country and category. Larger rectangles indicate more frequent object classes in the dataset, such as swimming pools, stadiums, and roads. The spatial diversity across countries, like the prevalence of crop fields in France and Italy, or urban structures in the United States and the Russian Federation, highlights the broad geographic and semantic coverage, which is critical for robust change analysis.}

    \label{fig:treemap}
\end{figure*}

\section{Methodology}

Our approach utilizes two techniques, one based on conversational VLMs and the other based on video-based VLMs, each optimized for different functionality necessary for our task. Conversational VLMs are fine-tuned for interactive conversations, remembering image features, and hence enabling dynamic comparison and support for "what-if" situations raised by users, thus making them \emph{interactive}. On the other hand, Video-based VLMs are optimized to track temporal variation between a pair of images, which facilitates detailed descriptions for each individual image and its variation. We utilized Video-based VLMs for temporal understanding comparison across them and our conversational VLM ChatENV.

\textbf{Qwen VLM:} We train ChatENV on the \textit{Qwen2.5-VL-7B-Instruct} model, which combines a Vision Transformer (ViT) image encoder together with a large-language model decoder~\cite{Qwen2.5}. This model is a multi-modal version of the Qwen 2.5 model, which is an open-source, light-weight model that is easy to finetune, with competitive performance on VQA and reasoning benchmarks. As shown in Figure~\ref{fig:overview}, the ViT, text encoder, and projection layer are left \emph{frozen}. Fine-tuning is restricted to light-weight Low-Rank Adaptation (LoRA) adapters placed within every attention and feedforward block within the decoder. A linear probe is optionally added for scalar prediction tasks. A training sample is an RGB aerial tile and a user prompt augmented with structured sensor data. Because environmental sensor readings (e.g., temperature, CO) cannot be directly inferred from an image, we provide them \textbf{\textit{explicitly}} as user input.

ChatENV can support three complementary tasks. In the \emph{single-turn} scenario, each sample features one image and an instruction such as: ``Describe this image''  and the model must produce a single ground-truth caption. In the \emph{two-turn what-if} scenario, the model captions an image and then answers a hypothetical ``What would happen if\ldots?'' question.  In the \emph{three-turn difference} scenario, the model captions two images sequentially and then responds to a ``What is the difference?'' prompt. We sample uniformly over pure captioning, comparative reasoning, and hypothetical tasks and train on cross-entropy loss over every assistant turn.

\textbf{Video-Based VLMs:} We employ two recent SOTA models: \textit{LLaVA-NeXT-Video}~\cite{liu2024llavanext} and \textit{Video-LLaVA}~\cite{liu2024videollava}. The \textit{LLaVA-NeXT-Video}, a video-adapted variant of LLaVA-NeXT, leverages advanced visual reasoning and improved general knowledge to provide accurate frame-level descriptions and reliable temporal comparisons, particularly excelling in zero-shot video understanding scenarios. Similarly, \textit{Video-LLaVA} effectively integrates temporal visual information with language understanding, achieving superior performance on video-language benchmarks and making it ideal for articulating subtle visual changes.

During fine-tuning, we convert every pair of images to a 2-frame video sequence within which every image is a separate frame. The frames are uniformly sampled by a video encoder to create visual embeddings, and in parallel, the text prompts are encoded. The visual and text features are optimized for alignment during training, which enhances those models' ability to describe every image separately and accurately find the fine-grained differences. We chose the 7B-parameter model variants to fit on one GPU for practical considerations.

\section{Experimental Setup}
ChatENV is built on the Qwen2.5-VL-7B-Instruct backbone and fine-tuned using LoRA with $r=64$ and $\alpha=128$, resulting in 190.36M trainable parameters (2.24\% of the full model). Experiments were conducted using BF16 precision on an NVIDIA RTX 5000 Ada GPU. Table~\ref{tab:runtime} reports inference latency, generation throughput, and memory usage across different task settings.

\textbf{LoRA}  
We fine-tuned our model using LoRA \cite{lora}. LoRA posits that the pre-trained and fine-tuned weights differ by a low-rank difference. Given a pre-trained matrix \(W_0 \in \mathbb{R}^{d \times k}\), LoRA decomposes the update as  
$W_0 + \Delta W = W_0 + BA$,  
where \(B \in \mathbb{R}^{d \times r}\), \(A \in \mathbb{R}^{r \times k}\), and \(r \ll \min(d,k)\). Here, \(W_0\) remains fixed, while \(A\) and \(B\) are trainable. The forward pass becomes  
$ h = W_0x + BAx$,  
scaled by \(\alpha / r\), with \(\alpha\) controlling the update magnitude. Typically, \(A\) is randomly initialized and \(B\) is zero-initialized, ensuring \(\Delta W\) starts at zero. For all the models and experiments, we set the rank (r) to 64, scale ($\alpha$) to 128, and dropout to 0.0, similar to the best results in \cite{elgendy2024geollava}.

\begin{table}[t]
\centering
\caption{Computational complexity and runtime efficiency of ChatENV.}
\label{tab:runtime}
\begin{tabular*}{\columnwidth}{@{\extracolsep{\fill}}lc}
\hline
\textbf{Metric} & \textbf{Value} \\
\hline
Model & Qwen2.5-VL-7B-Instruct \\
Total Parameters & 8.48B \\
Trainable Parameters & 190.36M (2.24\%) \\
LoRA $(r, \alpha)$ & $(64, 128)$ \\
Precision & BF16 \\
GPU & NVIDIA RTX 5000 Ada \\
\hline
Avg. Input Tokens (1 image / 2 images) & 1369 / 2747 \\
Avg. Generated Tokens & 365 \\
Avg. Seconds / Query & 10.60 \\
Avg. Tokens / Second & 26.83 \\
Peak VRAM (GB) & 17.07 \\
\hline
\end{tabular*}
\end{table}

\textbf{Evaluation Metrics:}
We take five metrics for strongly assessing generation quality, which capture both semantic similarity and surface-level overlap, and provide complementary insight into reference text alignment.

\textit{(I) ROUGE} \cite{lin2004rouge} computes $n$-gram recall between the generated and reference text. We report ROUGE-L, which rewards the longest common subsequence (LCS) and hence fluency and phrase-level overlap. ROUGE-1 and 2 evaluate unigram and bigram recall, reflecting lexical overlap. ROUGE is particularly well-suited for content coverage estimation.

\textit{(II) SBERT Similarity}
uses Sentence-BERT to encode generated and reference text into high-dimensional vectors, compared against each other using cosine similarity \cite{reimers2019sentence}. It detects semantic overlap even on rephrased sentences.

\textit{(III) BERTScore} \cite{zhang2019bertscore} is computed by comparing generated and reference tokens using contextual embeddings. We use DeBERTa V3 to calculate precision, recall, and F1, which reflect fluency and semantic similarity across different phrasings.

\textit{(IV) COMET} \cite{rei2020comet} is a neural metric that has been trained over human judgments. It projects source, hypothesis, and reference onto a common space and makes predictions for quality ratings for adequacy and fluency. COMET is especially robust for generation tasks and correlates strongly with human evaluations.

\textit{(V) Keyword Cluster Evaluation (KCE)} is a metric we introduce to evaluate directional language. Words such as \textit{rise}, \textit{increase}, \textit{fall}, and \textit{decrease} are essential for comparative descriptions, yet they are often overlooked or incorrectly handled by standard metrics. We extract common directional terms from ``difference'' sentences, group their variants into clusters, and compute F1-scores (KCE-F1) based on their presence in both the reference and hypothesis. This captures the model’s ability to detect and preserve perceptual changes in meaning.

The ground-truth annotations are: (i) an image description, (ii) a description of image differences, (iii) a \textit{what-if} question based on the differences, and (iv) an answer based on the second image. Example: if the first image has fewer buildings, increased wind speed, and increased temperature, then a sample question could be: \textit{``What if more buildings were present?''} A sample reference answer could be: \textit{``Wind speed might slow due to building-induced drag, and temperatures might increase due to more concrete surfaces.''} During what-if inference, the model produces an answer based on only the first image and the what-if scenario, \textbf{\textit{without seeing the second image}}. Model output is tested against reference answers using the same test data for every configuration.

All models were trained on a 48GB A6000 GPU for one epoch on the entire dataset. Further tests on ChatGPT and Gemini-specific annotations were done separately. For VLMs based on videos, a sample split equal to roughly 10\% of the dataset was built separately for ChatGPT and Gemini and used to test the data size effect on the models' performance.

\section{Results \& Discussion}

\begin{table*}[ht]
\centering
\caption{Model Performance on ChatENV (Qwen based), Best per column underlined, best per group colored.}
\label{tab:combined_results}
\begin{minipage}{0.54\textwidth}
\centering
\subcaption{Three-turn setting }
\label{tab:qwen_results}
\scriptsize
\resizebox{\textwidth}{!}{%
\begin{tabular}{@{}ll
                S[table-format=1.3]
                S[table-format=1.3]
                S[table-format=1.3]
                S[table-format=1.3]
                S[table-format=1.3]@{}}
\toprule
\textbf{Annotations} & \textbf{Training} & \textbf{ROUGE-L} & \textbf{SBERT} & \textbf{BERT-F1} & \textbf{COMET} & \textbf{KCE-F1} \\
\midrule
\cellcolor{gray!10} ChatGPT
  & Base          & 0.124 & 0.430 & 0.824 & 0.496 & 0.710 \\
  & LoRA          & \color{blue}0.242 & \color{blue}0.702 & \color{blue}0.889 & \color{blue}0.733 & \color{blue}0.818 \\
  & Lin. PROBE  & 0.233 & 0.648 & 0.884 & 0.699 & 0.817 \\
\midrule
  \cellcolor{gray!10}Gemini
  & Base          & 0.122 & 0.450 & 0.825 & 0.490 & 0.692 \\
  & LoRA          & \color{orange}\underline{0.298} & \color{orange}\underline{0.803} & \color{orange}\underline{0.902} & \color{orange}\underline{0.763} & 0.826 \\
  & Lin. PROBE  & 0.289 & 0.794 & 0.899 & 0.752 & \color{orange}\underline{0.830} \\
\midrule
\cellcolor{gray!10}ChatGPT$+$Gemini
  & Base          & 0.124 & 0.445 & 0.825 & 0.495 & 0.705 \\
  & LoRA          & \color{green}0.250 & \color{green}{0.737} & \color{green}0.890 & \color{green}{0.745} & \color{green}{0.814} \\
  & Lin. PROBE  & 0.236 & 0.706 & 0.883 & 0.713 & 0.809 \\
\bottomrule
\end{tabular}}
\end{minipage}%
\hfill
\begin{minipage}{0.42\textwidth}
\centering
\subcaption{Two-turn (What-If) setting}
\label{tab:whatif_results}
\scriptsize
\resizebox{\textwidth}{!}{%
\begin{tabular}{@{}l
                S[table-format=1.3]
                S[table-format=1.3]
                S[table-format=1.3]
                S[table-format=1.3]
                S[table-format=1.3]@{}}
\toprule
\textbf{Training} & \textbf{ROUGE-L} & \textbf{SBERT} & \textbf{BERT-F1} & \textbf{COMET} & \textbf{KCE-F1} \\
\midrule
Base          & 0.108 & \color{blue}0.607 & 0.840 & 0.627 & 0.587 \\
LoRA          & 0.231 & 0.597 & \color{blue}0.893 & \color{blue}0.686 & 0.800 \\
Lin. PROBE  & \color{blue}0.233 & 0.596 & 0.889 & 0.684 & \color{blue}0.813 \\
\midrule
Base          & 0.115 & 0.597 & 0.837 & 0.626 & 0.569 \\
LoRA          & \color{orange}\underline{0.282} & \color{orange}\underline{0.667} & \color{orange}\underline{0.900} & \color{orange}\underline{0.705} & \color{orange}\underline{0.816} \\
Lin. PROBE  & 0.247 & 0.647 & 0.889 & 0.702 & 0.759 \\
\midrule
Base          & 0.111 & 0.603 & 0.838 & 0.628 & 0.576 \\
LoRA          & \color{green}0.254 & \color{green}0.646 & \color{green}0.895 & \color{green}0.695 & \color{green}0.809 \\ Lin. PROBE  & 0.234 & 0.629 & 0.886 & 0.666 & 0.785 \\
\bottomrule
\end{tabular}}
\end{minipage}
\end{table*}

\textbf{Three-Turn Evaluation.} Table~\ref{tab:qwen_results} shows ChatENV’s performance on a three-turn task that (i) describes Image~1, (ii) describes Image~2, and (iii) explains their differences. Training is conducted using three annotation regimes, ChatGPT, Gemini (each covering 50\% of the data), and their union, and evaluation is done over a held-out, fixed test set comprising both sources' annotations. Additionally, sensor information are passed to the model for each image pair as part of the input prompt. All models share the same Qwen-2.5 backbone and decoder hyperparameters; the only difference lies in the adaptation strategy.

\textit{Zero-shot Reference:}
The base model is evaluated by directly passing ground-truth annotations without any fine-tuning, serving as a zero-shot baseline. As expected, performance is fair across most metrics, primarily due to the base model’s exposure to satellite imagery during pretraining. However, finetuning yields consistent improvements across all metrics.

\textit{LoRA \emph{Vs} Linear Probing:} Consistently injecting LoRA adapters enhances performance on every measurement relative to the baseline for both the individual and the combined annotation conditions. Importantly, for the combined set, LoRA increases the scores for SBERT and COMET by about $+0.30$. Linear probing, however, is competitive and close metrics compared to LoRA, and could potentially surpass it for KCE-F1 scores. Looking through individual sets, LoRA outperforms the probing method and reaches the best score in all five metrics for the ChatGPT-only condition. This indicates that LoRA improves semantic match and contextual understanding, while linear probing can prove to be an effective lightweight option for situations when surface similarity is most important or when resources are scarce.

\textbf{Two-Turn ``What-If'' Evaluation.}
Table~\ref{tab:whatif_results} shows {\it What-If} performance when the model is fed a single image and sensor data, generates a weather-aware caption, and then responds to a hypothetical question (e.g., ``What if more trees are planted?'') by predicting environmental changes like temperature increase or \textsc{CO\textsubscript{2}} reduction. Scores are generally consistent with the trend seen in Table~\ref{tab:qwen_results} but are typically lower in some metrics, since the model has to {\it imagine} the situation as there is no second image, making the task inherently more difficult.
In spite of this increased complexity, the \textsc{KEC-F1} and SBERT scores are still high, which verifies that the model identifies the direction and meaning of important changes accurately, which is the overall aim for the What-If scenario.

\begin{table*}[htbp]
\caption{Comparison between Video-LLaVA (left | ) and LLaVA-NeXT-Video ( | right). Best result per model per metric is colored. Best result per metric across all experiments is underlined.}
\centering
\renewcommand{\arraystretch}{1.2} 
\setlength{\tabcolsep}{5pt} 
\resizebox{\textwidth}{!}{%
\begin{tabular}{
>{\centering\arraybackslash}p{2.4cm} 
>{\centering\arraybackslash}p{1.3cm} 
>{\centering\arraybackslash}p{1.6cm}@{}!{\vrule width 0.4pt}@{}>{\centering\arraybackslash}p{1.6cm} 
>{\centering\arraybackslash}p{1.6cm}@{}!{\vrule width 0.4pt}@{}>{\centering\arraybackslash}p{1.6cm} 
>{\centering\arraybackslash}p{1.6cm}@{}!{\vrule width 0.4pt}@{}>{\centering\arraybackslash}p{1.6cm} 
>{\centering\arraybackslash}p{1.6cm}@{}!{\vrule width 0.4pt}@{}>{\centering\arraybackslash}p{1.6cm} 
>{\centering\arraybackslash}p{1.6cm}@{}!{\vrule width 0.4pt}@{}>{\centering\arraybackslash}p{1.6cm} 
}
\toprule
\textbf{Annotations} & \textbf{Training} & \multicolumn{2}{c}{\textbf{ROUGE-L}} & \multicolumn{2}{c}{\textbf{SBERT}} & \multicolumn{2}{c}{\textbf{BERT-F1}} & \multicolumn{2}{c}{\textbf{COMET}} & \multicolumn{2}{c}{\textbf{KCE-F1}} \\
\midrule
\cellcolor{gray!10}ChatGPT
& Base  & 0.116 & 0.118 & 0.562 & 0.563 & 0.540 & 0.546 & 0.436 & 0.444 & 0.495 & 0.500 \\
& 10K   & \textcolor{blue}{0.255} & 0.172 & 0.826 & 0.696 & \textcolor{blue}{0.757} & 0.674 & 0.785 & 0.641 & 0.709 & 0.649 \\
& 76K   & 0.224 & \textcolor{blue}{\underline{0.243}} & \textcolor{blue}{0.859} & \textcolor{blue}{0.868} & 0.743 & \textcolor{blue}{0.776} & \textcolor{blue}{0.789} & \textcolor{blue}{0.800} & \textcolor{blue}{0.736} & \textcolor{blue}{\underline{0.758}} \\
& Lin. Probe & 0.147 & 0.206 & 0.675 & 0.767 & 0.634 & 0.731 & 0.647 & 0.761 & 0.622 & 0.652 \\
\midrule
\cellcolor{gray!10}Gemini
& Base  & 0.073 & 0.079 & 0.551 & 0.552 & 0.512 & 0.519 & 0.414 & 0.421 & 0.450 & 0.455 \\
& 10K & \textcolor{orange}{\underline{0.299}} & \textcolor{orange}{\underline{0.243}} & \textcolor{orange}{\underline{0.897}} & \textcolor{orange}{\underline{0.896}} & \textcolor{orange}{\underline{0.818}} & \textcolor{orange}{\underline{0.804}} & \textcolor{orange}{\underline{0.833}} & \textcolor{orange}{\underline{0.831}} & \textcolor{orange}{0.729} & 0.717 \\
& 76K   & 0.246 & 0.238 & 0.868 & 0.894 & 0.774 & 0.770 & 0.805 & 0.824 & 0.720 & \textcolor{orange}{0.732} \\
& Lin. Probe & 0.194 & 0.242 & 0.783 & 0.819 & 0.694 & 0.754 & 0.716 & 0.768 & 0.622 & 0.636 \\
\midrule
\cellcolor{gray!10}\mbox{ChatGPT+Gemini}
& Base  & 0.095 & 0.099 & 0.557 & 0.558 & 0.526 & 0.553 & 0.425 & 0.433 & 0.472 & 0.478 \\
& 15K   & \textcolor{green}{0.238} & \textcolor{green}{0.208} & \textcolor{green}{0.861} & \textcolor{green}{0.820} & 0.770 & 0.755 & 0.810 & 0.806 & 0.639 & 0.688 \\
& 152K  & 0.234 & 0.205 & 0.849 & 0.809 & \textcolor{green}{0.784} & \textcolor{green}{0.769} & \textcolor{green}{0.816} & \textcolor{green}{0.812} & \textcolor{green}{\underline{0.757}} & \textcolor{green}{0.704} \\
& Lin. Probe & 0.215 & 0.201 & 0.825 & 0.705 & 0.735 & 0.761 & 0.748 & 0.798 & 0.626 & 0.685 \\
\bottomrule
\end{tabular}
}
\label{tab:video_results}
\end{table*}

\textbf{Comparison with SOTA Video VLMs.}
Table~\ref{tab:video_results} shows an ablation study comparing ChatENV and two top video models, Video-LLaVA and LLaVA-NeXT-Video architectures specifically tailored for temporal understanding. The two models treat the two input frames as a brief clip and are trained on a three-turn setup, which allows them to memorize directly. This achieves a modest advantage over most aggregated metrics; for example, the best video configuration achieves a COMET of 0.833 compared to 0.763 for ChatENV.

Most importantly, ChatENV still performs and can surpass the video models on certain metrics. It achieves the best BERT-F1 (0.902 vs.\ 0.818) and KCE-F1 (0.830 vs.\ 0.758) scores and almost the same ROUGE-L score, which reflects better agreement with human-annotated data despite needing to deduce temporal change under the absence of a sequential video stream.

Base (non-tuned) versions of the video models underperform the ChatENV baseline, emphasizing the role played by domain exposure: Qwen's pre-training corpus already contains satellite data, while Video-LLaVA and LLaVA-NeXT don't have aerial data. Upon fine-tuning with our curated dataset, video-model performance increases by at least 75\%, demonstrating the way the suggested data pipeline overcomes this domain gap.

From Tables \ref{tab:combined_results} and \ref{tab:video_results}, it is clear that Gemini annotations consistently beat out ChatGPT annotations on virtually all metrics. Comparing the two models from Table \ref{tab:video_results} further, Video-LLaVA tends to surpass LLaVA-NeXT-Video on ROUGE-L, COMET, and KCE-F1, among others, using fewer samples, for which Video-LLaVA, together with the Gemini annotations, achieves the best overall score.
\begin{figure*}[t!]
    \centering
    \includegraphics[width=0.90\textwidth]{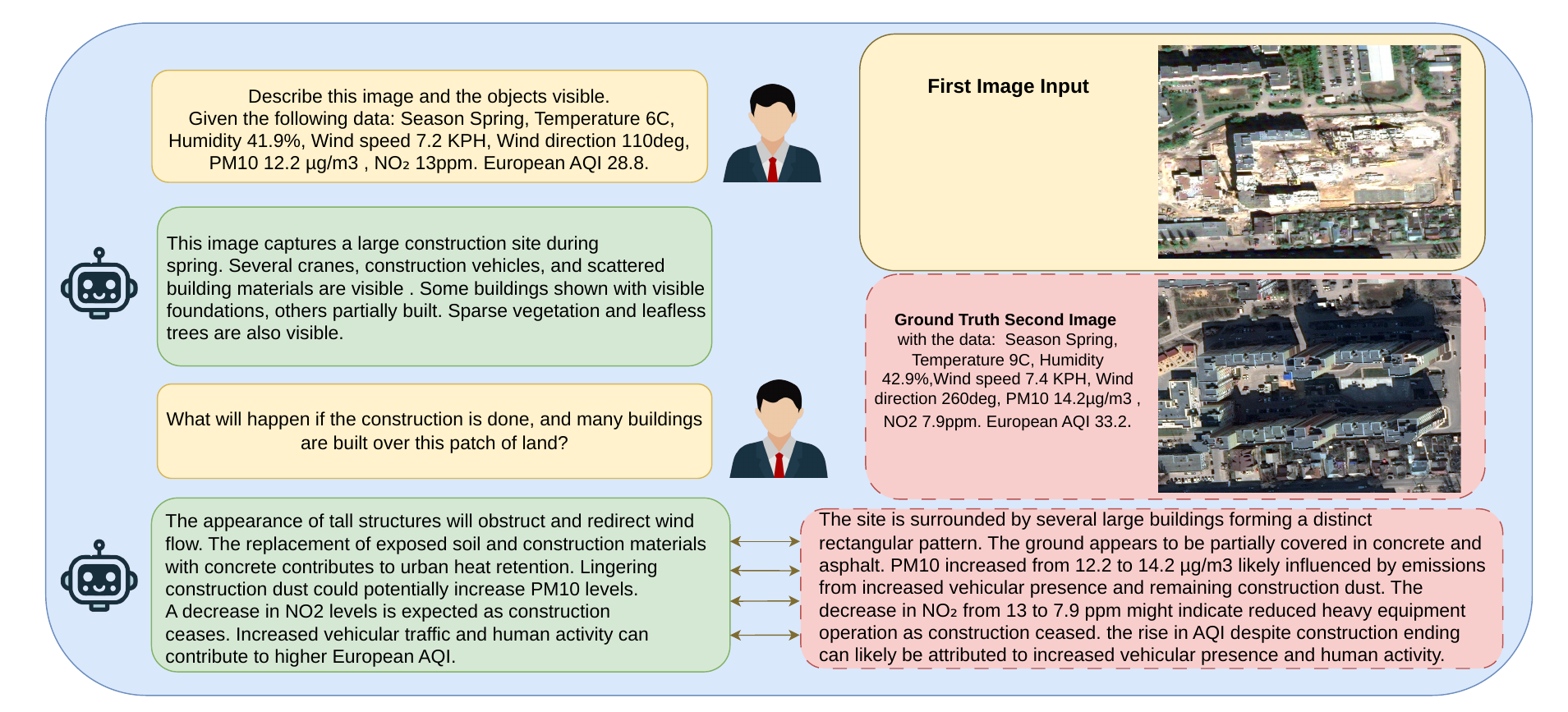}
    \caption{The figure illustrates a what-if interaction with ChatENV. Given the initial image and environmental metadata, the user poses a scenario question: \emph{“What will happen if construction is done, and many buildings are built over this patch of land?”}The model generates a detailed answer that closely matches the second (ground-truth) image’s description, such as increased PM10 due to dust and traffic, decreased $NO_2$ following construction, altered wind patterns, and heat retention from concrete.}
    \label{fig:visual_results}
\end{figure*}

Table \ref{tab:ablation-sensors} (LEFT) showcases the effect of having the sensor information in the prompt versus not having them. This experiment was done using the Three-Turn Setting for the ChatENV model, using LoRA finetuning. It is clear that for all the annotation sets, including the sensor data yields higher scores in all metrics, demonstrating the need for sensor data to further explain and describe certain scenes in the images.

\begin{table*}[t]
\centering
\caption{Ablation study under the LoRA (three-turn) setting for the ChatENV model. Left: With vs. without sensor data. Right: Comparison across different vision-language models.}
\label{tab:ablation-full}
\scriptsize
\renewcommand{\arraystretch}{0.9}
\setlength{\tabcolsep}{3pt}
\begin{minipage}{0.48\textwidth}
\centering
\subcaption{With vs. without sensor data}
\label{tab:ablation-sensors}
\resizebox{\textwidth}{!}{%
\begin{tabular}{@{}l l r r r r r@{}}
\toprule
\textbf{Annotations} & \textbf{Sensor} & \textbf{ROUGE-L} & \textbf{SBERT} & \textbf{BERT-F1} & \textbf{COMET} & \textbf{KCE-F1} \\
\midrule
\cellcolor{gray!10}ChatGPT
  & w/o & 0.232 & 0.691 & 0.888 & 0.721 & 0.809 \\
  & w/  & \textbf{0.242} & \textbf{0.702} & \textbf{0.889} & \textbf{0.733} & \textbf{0.818} \\
  & $\Delta$ & +4.3\% & +1.6\% & +0.1\% & +1.7\% & +1.1\% \\
\midrule
\cellcolor{gray!10}Gemini
  & w/o & 0.291 & 0.799 & 0.901 & 0.762 & 0.815 \\
  & w/  & \textbf{0.298} & \textbf{0.803} & \textbf{0.902} & \textbf{0.763} & \textbf{0.826} \\
  & $\Delta$ & +2.4\% & +0.5\% & +0.1\% & +0.1\% & +1.4\% \\
\midrule
\cellcolor{gray!10}\raisebox{-10pt}[0pt][0pt]{\shortstack{ChatGPT\\+Gemini}}
  & w/o & 0.222 & 0.684 & 0.887 & 0.724 & 0.802 \\
  & w/  & \textbf{0.250} & \textbf{0.737} & \textbf{0.890} & \textbf{0.745} & \textbf{0.814} \\
  & $\Delta$ & +12.6\% & +7.7\% & +0.3\% & +2.9\% & +1.5\% \\
\bottomrule
\end{tabular}}
\end{minipage}
\hfill
\begin{minipage}{0.49\textwidth}
\centering
\subcaption{Comparison with other change detection models}
\label{tab:ablation-models}
\resizebox{\textwidth}{!}{%
\begin{tabular}{@{}l l r r r r r@{}}
\toprule
\textbf{Annotations} & \textbf{Model} & \textbf{ROUGE-L} & \textbf{SBERT} & \textbf{BERT-F1} & \textbf{COMET} & \textbf{KCE-F1} \\
\midrule
\cellcolor{gray!10}ChatGPT
  & TEOChat    & 0.128 & 0.485 & 0.511 & 0.506 & 0.433 \\
  & RS-LLaVA   & 0.180 & 0.655 & 0.611 & 0.638 & 0.530 \\
  & \textbf{ChatENV} & \textbf{0.242} & \textbf{0.702} & \textbf{0.889} & \textbf{0.733} & \textbf{0.818} \\
\midrule
\cellcolor{gray!10}Gemini
  & TEOChat    & 0.114 & 0.453 & 0.520 & 0.499 & 0.398 \\
  & RS-LLaVA   & 0.155 & 0.621 & 0.590 & 0.611 & 0.493 \\
  & \textbf{ChatENV} & \textbf{0.298} & \textbf{0.803} & \textbf{0.902} & \textbf{0.763} & \textbf{0.826} \\
\midrule
\cellcolor{gray!10}\raisebox{-10pt}[0pt][0pt]{\shortstack{ChatGPT\\+Gemini}}
  & TEOChat    & 0.135 & 0.498 & 0.531 & 0.525 & 0.421 \\
  & RS-LLaVA   & 0.161 & 0.621 & 0.588 & 0.618 & 0.511 \\
  & \textbf{ChatENV} & \textbf{0.250} & \textbf{0.737} & \textbf{0.890} & \textbf{0.745} & \textbf{0.814} \\
\bottomrule
\end{tabular}}
\end{minipage}
\end{table*}

Figure \ref{fig:visual_results} provides a qualitative result of the \textit{What-If} interaction flow. A user first queries the model about an aerial image with its associated sensor readings, then poses a hypothetical (``What if \ldots?'') question. ChatENV’s response (green) closely matches the reference answer derived from the second image (red), where the what-if scenario answer actually happens (using the second image in the dataset). This demonstrates its ability to reason about unseen outcomes. This fidelity underpins practical use-cases such as environmental monitoring and urban-planning simulations, where grounded yet imaginative projections are essential.

\begin{figure*}[t!]
    \centering
    \includegraphics[width=0.85\textwidth]{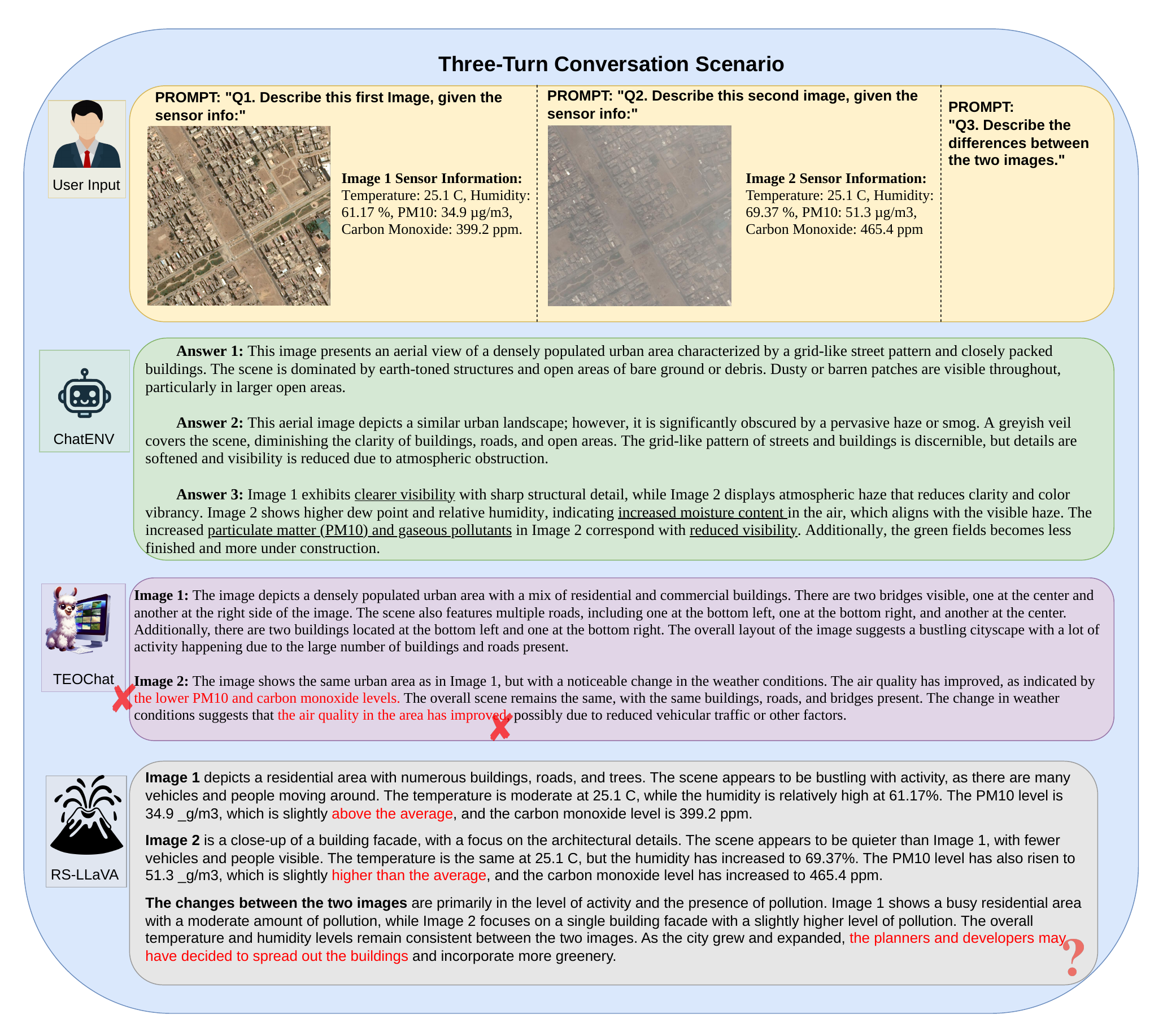}
    \caption{Three-turn qualitative comparison with sensor-conditioned prompts. We keep question intent fixed but adapt prompt formatting to each model’s supported interface. ChatENV coherently grounds scene-level changes using paired sensor context, while baseline errors are highlighted (red \redx: sensor inconsistency/misinterpretation; \redq: speculative/weakly grounded claim).}
    \label{fig:three_turn_output}
\end{figure*}

\textbf{Qualitative Comparison with SOTA Temporal/Change VLMs:}  We qualitatively compare ChatENV against two recent change detection remote sensing VLMs, TEOChat and RS-LLaVA, using a three-turn sensor-paired dialogue (Figure ~\ref{fig:three_turn_output}). The user sequentially requests (Q1) a description of the first image, (Q2) a description of the second image, and (Q3) an explanation of the differences, while providing the paired environmental readings for each timestamp. ChatENV produces consistent scene-level descriptions across turns and coherently links the reduced visibility in the second image to the concurrent changes in atmospheric conditions reflected by the sensor readings. In contrast, although TEOChat can ingest bi-temporal inputs, it frequently confuses or misinterprets the associated sensor values, leading to incorrect conclusions (e.g., claiming improved air quality when pollutant levels increase). RS-LLaVA processes images sequentially, but it tends to treat sensor values as ungrounded metadata, often reporting only value trends (increase/decrease) without connecting them to the observed visual changes. These qualitative behaviors align with the quantitative results in Table~\ref{tab:ablation-models}, where ChatENV achieves higher scores across all metrics.

\section{Conclusion}
This paper introduces ChatENV, a novel interactive environmental VLM that jointly infers over satellite images and real-world sensor data. Through the creation of the largest multimodal dataset of its kind, combining pairs of aerial images and rich environmental metadata, with heterogeneous dual-model annotations, ChatENV enables comprehensive and context-aware environmental understanding. The model facilitates fine-grained descriptions for scenes, spatiotemporal comparisons, and “what-if” scenario simulation. Experimental results show ChatENV could output meaningful and humanized descriptions that are very important for urban planning and environmental monitoring.

\noindent\textbf{Limitations and Future Work:}
While ChatENV shows that combining temporal imagery with environmental context improves interactive monitoring and scenario-style querying, limitations remain. First, third-party API measurements can be geographically uneven or missing. Second, point-based readings may not match the spatial footprint of an image tile. Third, the current benchmark focuses on bi-temporal pairs; extending it to longer time series could enable stronger trend reasoning. Future work will (i) integrate additional modalities (e.g., SAR, multispectral, socio-economic data), (ii) incorporate memory-augmented agents for long-term reasoning, (iii) support real-time deployment via continual LoRA updates, and (iv) explore joint video–dialogue pretraining. These steps position ChatENV toward becoming a foundation model for predictive environmental intelligence.


\bibliographystyle{IEEEtran}
\bibliography{egbib}

\begin{IEEEbiography}[{\includegraphics[width=1.25in,height=1.25in,clip,keepaspectratio]{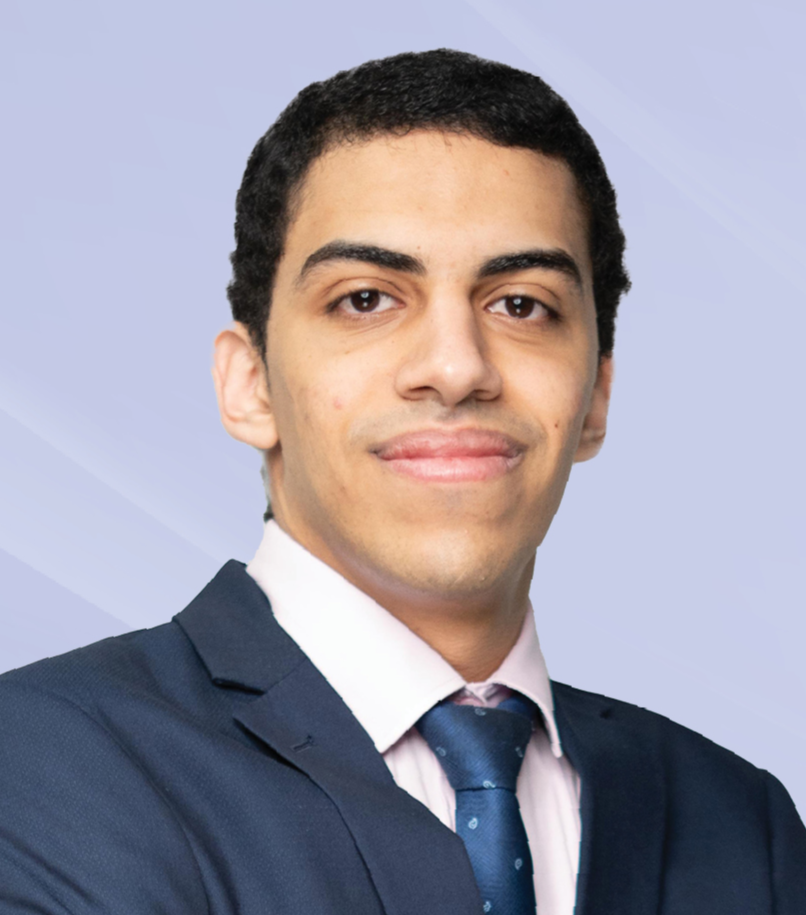}}]{Hosam Elgendy} is currently a Research Engineer in the Machine Learning department at Mohamed bin Zayed University of Artificial Intelligence (MBZUAI) in Abu Dhabi, UAE. He previously received his Master's degree in Computer Vision from MBZUAI in 2024. His research focuses on vision–-language models and multi--modal learning systems and his interests span computer vision, natural language processing, and multi-modal data fusion.
\end{IEEEbiography}

\begin{IEEEbiography}
[{\includegraphics[width=1in,height=1.25in,clip,keepaspectratio]{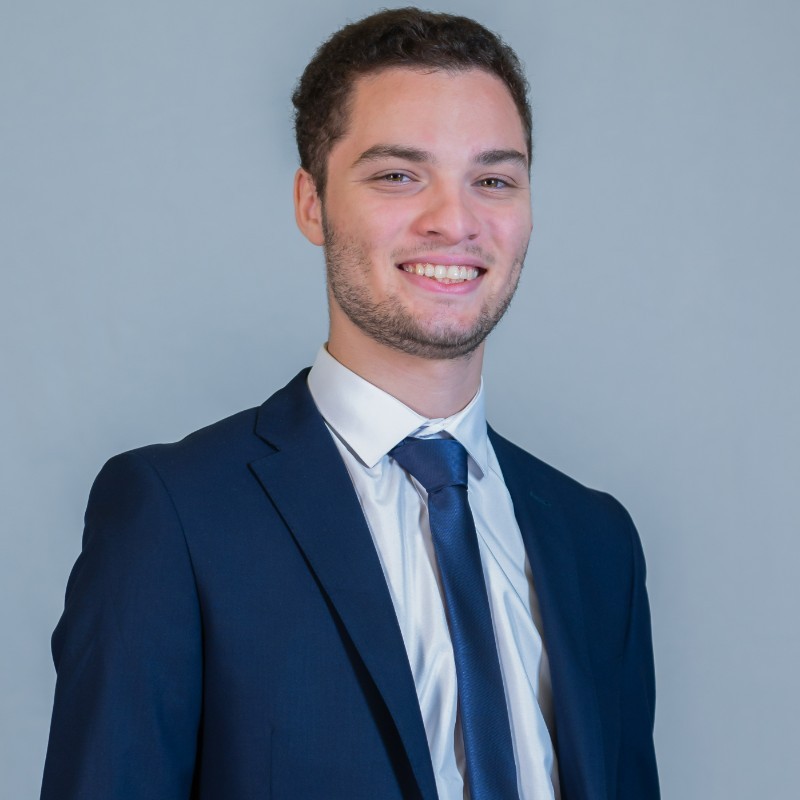}}]{Ahmed Sharshar} is currently pursuing a PhD in Computer Vision at MBZUAI in Abu Dhabi, UAE. He previously received his Master's degree in Computer Vision from MBZUAI. He obtained his Bachelor of Engineering degree in Computer Engineering from the Egypt-Japan University of Science and Technology (E-JUST), Egypt.
His research primarily focuses on developing lightweight models and expanding their applications across various domains, such as natural language processing, computer vision, and human-computer interaction.
\end{IEEEbiography}

\begin{IEEEbiography}
[{\includegraphics[width=1in,height=1.25in,clip,keepaspectratio]{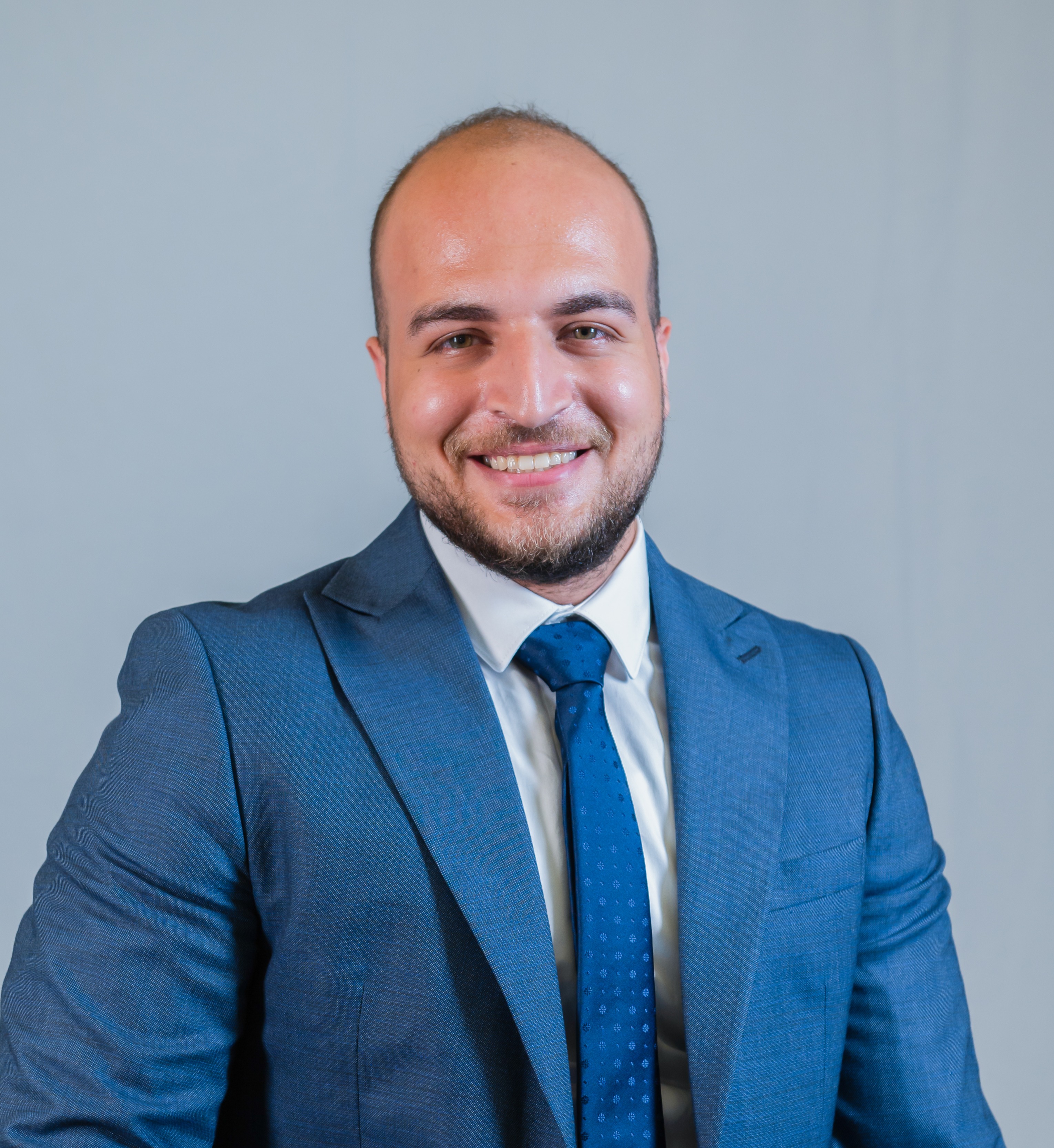}}] {Ahmed Hesham} is an AI researcher specializing in Natural Language Processing and speech technologies. He earned his Master’s degree in Natural Language Processing from MBZUAI in Abu Dhabi, UAE, and his Bachelor’s degree in Computer Engineering from E-JUST, Egypt. His research spans large language models, retrieval-augmented generation, and speech processing, with applications in healthcare, multilingual NLP, and Arabic language technologies.
\end{IEEEbiography}

 \begin{IEEEbiography}[{\includegraphics[width=1.5in,height=1.25in,clip,keepaspectratio]{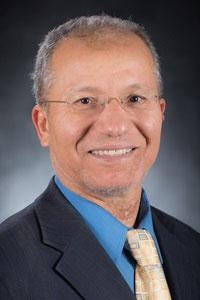}}]{Mohsen Guizani} (S’85, M'89, SM'99, F’09) received his B.S. (with distinction) and M.S. degrees in electrical engineering, and M.S. and Ph.D. degrees in computer engineering from Syracuse University, New York, in 1984, 1986, 1987, and 1990, respectively. He is currently a Professor with the Machine Learning Department, MBZUAI, Abu Dhabi, UAE. Previously, he served in different academic and administrative positions at the University of Idaho, Western Michigan University, the University of West Florida, the University of Missouri-Kansas City, the University of Colorado-Boulder, and Syracuse University. His research interests include wireless communications and mobile computing, computer networks, mobile cloud computing, security, and smart grid. He was the Editor-in-Chief of IEEE Network. He serves on the Editorial Boards of several international technical journals, and is the Founder and Editor-in-Chief of the Wireless Communications and Mobile Computing journal (Wiley). He is the author of nine books and more than 500 publications in refereed journals and conferences. He has guest edited a number of Special Issues in IEEE journals and magazines. He has also served as a TPC member, Chair, and General Chair of a number of international conferences.
\end{IEEEbiography}

\vfill

\newpage
\appendices
\onecolumn
\section{Annotation Details and Scoring Criteria}

These are the full details for scores based on the criteria to manually evaluate the GPT-4o and Gemini 2.0 annotations. These were presented to the annotators (authors) when reviewing the samples. Please note that the total set was split among the annotators, with no overlap due to the subjectivity behind rating the samples. This eliminates the inter-rater overlap.

\subsection*{Scoring Criteria}

\textbf{Question 1: Is each description accurate and captures details that are actually visible in the image?}

\begin{itemize}
    \item \textbf{1 Point:} The description does not capture any details in the image.
    \item \textbf{2 Points:} The description captures a few details but misses key visible elements/objects or includes inaccuracies.
    \item \textbf{3 Points:} The description captures most major details but omits minor but visible elements/objects.
    \item \textbf{4 Points:} The description is mostly accurate, including most of the key details, with only minimal omissions.
    \item \textbf{5 Points:} The description includes all objects visible in the image and describes details with high accuracy.
\end{itemize}

\vspace{\baselineskip}

\textbf{Question 2: Does the description include or describe the "class" of the images correctly?}

\begin{itemize}
    \item \textbf{1 Point:} The description entirely misidentifies or ignores the class label.
    \item \textbf{2 Points:} The description attempts to include the class but is incorrect or incomplete.
    \item \textbf{3 Points:} The description includes the class but with partial or unclear representation.
    \item \textbf{4 Points:} The description includes the class clearly but may lack context or supporting details.
    \item \textbf{5 Points:} The description accurately identifies the class and fully integrates within the image context.
\end{itemize}

\vspace{\baselineskip}

\textbf{Question 3: Does the model focus more on the majority or minority classes in the image?}

\begin{itemize}
    \item \textbf{1 Point:} The description fails to focus on either the majority or minority classes.
    \item \textbf{2 Points:} The description only focuses on the majority class and completely ignores minority classes.
    \item \textbf{3 Points:} The description prioritizes the majority class and gives minimal attention to minority classes.
    \item \textbf{4 Points:} The description gives balanced attention to both majority and minority classes but with insufficient detail.
    \item \textbf{5 Points:} The description provides equal and detailed attention to both majority and minority classes.
\end{itemize}

\subsection*{Sample Distribution}
Based on this grading, each question gets a score between 3 and 15. We chose to keep the samples with a score of 9 or above as mentioned in Table \ref{table:sample_distribution}.

\begin{table}[t!]
\centering
\begin{tabular}{|c|c|}
\hline
\textbf{Total Score} & \textbf{Number of Samples} \\ \hline
3                    & 86                         \\ \hline
4                    & 166                        \\ \hline
5                    & 245                        \\ \hline
6                    & 290                        \\ \hline
7                    & 224                        \\ \hline
8                    & 227                        \\ \hline
\textbf{Subtotal ($<$9) }     & \textbf{1,238}                      \\ \hline
9                    & 655                        \\ \hline
10                   & 1,004                      \\ \hline
11                   & 1,327                      \\ \hline
12                   & 1,659                      \\ \hline
13                   & 1,991                      \\ \hline
14                   & 1,825                      \\ \hline
15                   & 1,562                      \\ \hline
\textbf{Subtotal ($\geq9$)}   & \textbf{10,023}                    \\ \hline
\textbf{Total}       & \textbf{11,261}            \\ \hline
\end{tabular}
\vspace{1em}
\caption{Distribution of samples by total score.}
\label{table:sample_distribution}
\end{table}

\newpage
\section{Ablation Studies for Video Models}
Additional ablation studies for the Video-LLaVA and LLaVA-NeXT-Video models with and without sensor information summarized in table \ref{tab:ablation-sensors-extended}. Similar to the discussion in the Results section, the video models are trained to performed well on temporal tasks, but these models fall short in tasks with satellite images. However, unlike Qwen-based models that are already trained on satellite data, the addition of sensors to the prompts for the video models shows a greater improvement over the baseline without the sensors. 

\begin{table}[t]
\centering
\caption{Extended ablation study using 10K/15K samples under the LoRA setting for the Video-LLaVA and LLaVA-NeXT-Video models showing performance with and without sensor data. Bold numbers represent sensor-enabled scores.}
\label{tab:ablation-sensors-extended}
\scriptsize
\renewcommand{\arraystretch}{0.9}
\setlength{\tabcolsep}{3pt}
\resizebox{0.99\textwidth}{!}{%
\begin{tabular}{@{}l l
                S[table-format=1.3] S[table-format=1.3] S[table-format=1.3] S[table-format=1.3] S[table-format=1.3]
                S[table-format=1.3] S[table-format=1.3] S[table-format=1.3] S[table-format=1.3] S[table-format=1.3]@{}}
\toprule
\textbf{Annotations} & \textbf{Sensor} 
& \multicolumn{5}{c}{\textbf{Video-LLaVA}} 
& \multicolumn{5}{c}{\textbf{LLaVA-NeXT-Video}} \\
\cmidrule(lr){3-7} \cmidrule(lr){8-12}
& & \textbf{ROUGE-L} & \textbf{SBERT} & \textbf{BERT-F1} & \textbf{COMET} & \textbf{KCE-F1}
  & \textbf{ROUGE-L} & \textbf{SBERT} & \textbf{BERT-F1} & \textbf{COMET} & \textbf{KCE-F1} \\
\midrule
\cellcolor{gray!10}ChatGPT
  & w/o & 0.157 & 0.564 & 0.539 & 0.583 & 0.429 & 0.108 & 0.494 & 0.491 & 0.546 & 0.471 \\
  & w/  & \textbf{0.255} & \textbf{0.826} & \textbf{0.757} & \textbf{0.785} & \textbf{0.709} 
          & \textbf{0.172} & \textbf{0.696} & \textbf{0.674} & \textbf{0.641} & \textbf{0.649} \\
\midrule
\cellcolor{gray!10}Gemini
  & w/o & 0.146 & 0.521 & 0.551 & 0.575 & 0.394 & 0.108 & 0.471 & 0.546 & 0.546 & 0.445 \\
  & w/  & \textbf{0.299} & \textbf{0.897} & \textbf{0.818} & \textbf{0.833} & \textbf{0.729} 
          & \textbf{0.243} & \textbf{0.896} & \textbf{0.804} & \textbf{0.831} & \textbf{0.717} \\
\midrule
\cellcolor{gray!10}\raisebox{-10pt}[0pt][0pt]{\shortstack{ChatGPT\\+Gemini}}
  & w/o & 0.146 & 0.512 & 0.554 & 0.574 & 0.430 & 0.145 & 0.532 & 0.534 & 0.602 & 0.458 \\
  & w/  & \textbf{0.238} & \textbf{0.861} & \textbf{0.770} & \textbf{0.810} & \textbf{0.639}
          & \textbf{0.208} & \textbf{0.820} & \textbf{0.755} & \textbf{0.806} & \textbf{0.688} \\
\bottomrule
\end{tabular}}
\end{table}

\end{document}